\documentclass{Interspeech}

 \interspeechcameraready

\title{SSAVSV: Towards Unified Model for Self-Supervised Audio-Visual Speaker Verification}

\author[]{R. Gnana Praveen}{}
\author[]{Jahangir}{Alam}

\affiliation{Computer Research Institute of Montreal (CRIM), Canada}{}{}
\email{}

\usepackage{comment}

\usepackage{multirow}
\begin{document}

\maketitle

\begin{abstract}
Conventional audio-visual methods for speaker verification rely on large amounts of labeled data and separate modality-specific architectures, which is computationally expensive, limiting their scalability. To address these problems, we propose a self-supervised learning framework based on contrastive learning with asymmetric masking and masked data modeling to obtain robust audiovisual feature representations. In particular, we employ a unified framework for self-supervised audiovisual speaker verification using a single shared backbone for audio and visual inputs, leveraging the versatility of vision transformers. The proposed unified framework can handle audio, visual, or audiovisual inputs using a single shared vision transformer backbone during training and testing while being computationally efficient and robust to missing modalities. Extensive experiments demonstrate that our method achieves competitive performance without labeled data while reducing computational costs compared to traditional approaches.

\end{abstract}

\section{Introduction}
Speaker verification is the task of biometric authentication that verifies the identity of a person based on the speech signal by comparing them with pre-enrolled speaker templates. 
It has a wide range of applications, including user authentication, access control, secure verification, and forensic investigations. Another widely explored research paradigm for biometric authentication is face verification, which has also shown significant advancement in the computer vision community \cite{WANG2021215}. 
With the advancement of deep learning architectures \cite{Desplanques2020,liou24_interspeech,7298682}, novel loss functions \cite{8953658,8682749} and pre-training \cite{10.1007/978-3-031-19778-9_7,10096659}, both speaker and face verification systems have achieved impressive performance. Despite the remarkable success of individual face and speaker verification systems, their performance degrades dramatically in challenging environments \cite{8683477,9350195}. For example, speech signals can be corrupted by interference or noise such as laughter, music, and other sounds. Similarly, face images can be corrupted due to extreme pose, low resolution, and motion blur. To address this problem, audio-visual fusion has recently gained a lot of attention as they can offer complementary relationships, improving performance even when one of the modalities is corrupted or missing \cite{8683477}.   

Traditionally, audio-visual methods for speaker verification have been achieved using a simple fusion of scores obtained from individual face and speaker verification systems, which was found to outperform unimodal approaches \cite{251181}. To further improve performance, attention models have been explored to efficiently capture synergic relationships by exploiting the rich associations between audio and visual embeddings \cite{sun23_interspeech,praveen2023audiovisual}. 
However, the performance of these audiovisual methods is constrained by the quality and availability of labeled data, which is labor intensive and costly to obtain. Moreover, reliance on labeled data creates a bottleneck, restricting the scalability of these models. Self-Supervised Learning (SSL) has emerged as a promising research direction to learn meaningful information from abundant data without the need for labels \cite{10559458}. Although speech-based SSL methods have received significant attention \cite{kim24c_interspeech,fathan24_interspeech}, the prospect of SSL methods for audiovisual speaker verification still remains an under-explored problem. Despite the potential of SSL methods to enhance the scalability of models, one of the major challenges associated with audio-visual speaker verification is the intense computational complexity involved in employing separate modality-specific architectures.

Motivated by these limitations, we investigate the prospect of developing a unified audio-visual learning framework for self-supervised speaker verification while maintaining low computational complexity.
The benefits of the proposed self-supervised unified framework across modalities are multifold. First, it eliminates the need for hand-crafted priors and modality-specific inductive biases, allowing for data-driven representation learning with minimal manual effort. Second, a unified framework provides greater flexibility to handle the individual audio or visual inputs as well as audio-visual inputs, making it more robust to missing modalities. Third, unified models are parameter-efficient and support scalability to large-scale datasets and bigger models, enabling foundation models, which can be adapted to wide-range of downstream tasks of multiple modalities.

Recently, unified approaches have been gaining attention for audio-visual tasks such as event localization \cite{9964072} or speech recognition \cite{haliassos2024unified}. However, these approaches still employ modality-specific training strategies or partial sharing of model weights to deal with the heterogeneity across the audio and visual modalities. Vision Transformers (ViTs) have been shown to be able to generalize well to both audio and visual modalities by representing audio as a 2D spectrogram (2D spatial structure) \cite{LAVISH_CVPR2023,10.1007/978-3-031-72630-9_18}. Inspired by these works, we propose a modality-agnostic unified framework for speaker verification by sharing the full set of parameters to obtain audio and visual embeddings with self-supervised learning paradigm, providing a unified audio-visual framework with reduced computational complexity. Following the idea of \cite{gong2023contrastive}, we explore the complementary relationships between the audio-visual contrastive learning and masked data modeling in our self-supervised learning framework. 
To effectively capture cross-modal relationships, we introduce asymmetric masking with the contrastive audio-visual matching objective of our self-supervised learning framework.   
The major contributions of this work can be summarized as: (1) To the best of our knowledge, this is the first work to explore self-supervised learning framework for audio-visual speaker verification by leveraging the audio-visual contrastive learning with masked data modeling. (2) A unified audio-visual framework is proposed to obtain robust audio and visual feature representations, offering flexibility to adapt to missing modalities with low computational complexity. (3) Asymmetric masking is introduced to effectively capture audio-visual correspondences with the contrastive learning objective. (4) Extensive experiments are conducted on Voxceleb 1 dataset to demonstrate the effectiveness of the proposed approach. 

\begin{figure*}[t]
  \centering  \includegraphics[width=0.85\linewidth]{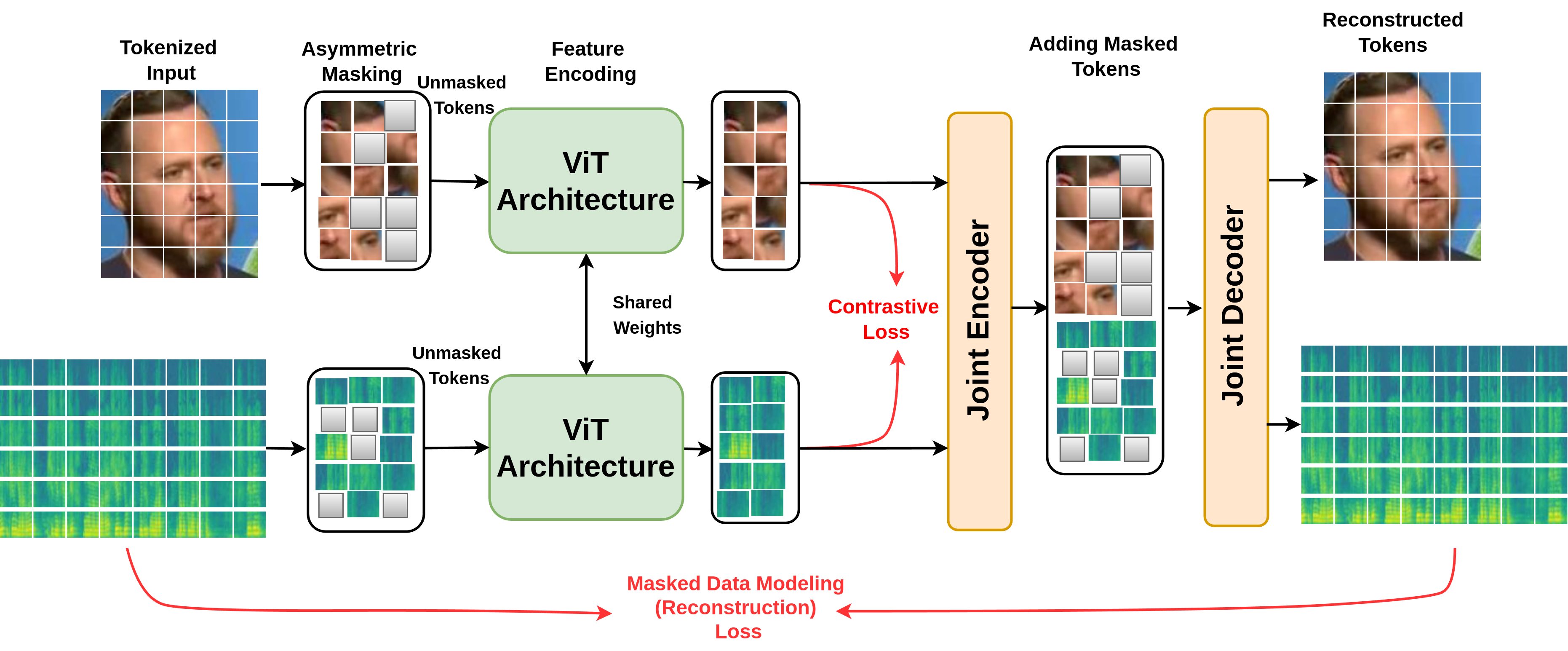}
  \caption{Block Diagram of the proposed approach}
  \label{fig:speech_production}
\end{figure*}

\section{Related Work}
The close association between faces and voices has been an active area of interest to mimic the human ability to integrate voice and face signals \cite{wen2021seeking,Nagrani18c}.
Wen et al. \cite{DBLP:conf/iclr/WenILRS19} proposed a disjoint mapping network by projecting the common covariates of the individual modalities into a shared representation space. The idea of projecting modality-specific features to a shared common representation space has become a defacto standard for several cross-modal processing tasks \cite{9414260,Tao2020}. Zhu et al. \cite{zhu2022unsupervised} explored self-supervision to capture the correspondences between faces and voices by enforcing relative inter-class separation within and across the modalities. 
Unlike these works, our work differs by using a single ViT backbone that shares the full set of parameters across the audio and visual modalities, while employing self-supervised learning objectives of audio-visual contrastive learning and masked data modeling. 

Sun et al. \cite{sun23_interspeech} investigated the relationship of keyframes between modalities and enhanced the saliency of keyframes using weight-enhanced attentive statistics pooling and joint attentive pooling, followed by gated attention. Selvakumar et al. \cite{selvakumar24_interspeech} showed that multi-task learning can improve the quality of audio-visual representations by employing an auxiliary task of age estimation, while leveraging multimodal augmentation to achieve better generalization. 
Tao et al. \cite{10096814} proposed a two-step deep cleansing framework to mitigate the impact of noisy labels by exploiting the complementary relationships across audio and visual modalities. Recently, cross-attention models have demonstrated significant performance improvements by utilizing synergistic relationships between audio and visual modalities \cite{praveen2024dynamic,praveen2024audiovisual,gnanapraveen24_odyssey}. However, these methods are based on supervised learning, whose performance is constrained by the availability of data. 

Cai et al. \cite{9741340} explored the potential of complementary information between audio and visual modalities to act as supervisory signal by using contrastive learning on audio data to generate pseudo-labels, followed by iterative clustering of audio and visual representations. Chen et al. \cite{10096925} further improved the idea of \cite{9741340} by incorporating co-meta learning to leverage the consistency between the modalities in the iterative clustering of audio and visual representations. Recently, Han et al. \cite{10314722} explored label correction methods to refine the pseudo-labels obtained from the self-distillation framework without labels. They further showed performance improvement by integrating visual information in iterative clustering stage to obtain more reliable pseudo-labels. These works used visual information as supplementary information in the framework of supervised learning to enhance audio-based speaker verification. Contrary to these works, we explicitly focus on leveraging the audio-visual correspondences using self-supervised learning paradigms of contrastive learning and masked data modeling.  



\section{Unified Self-Supervised A-V Model}

\subsection{Audio-Visual Inputs}
Given the video sequence, we extract audio and visual streams. The visual stream is pre-processed using RetinaFace \cite{9157330} to obtain the cropped and aligned facial images. For the visual modality, we use a randomly sampled image of size $H_v\times W_v\times 3$ from the preprocessed facial images, where $H_v$ and $W_v$ denote the height and width of the sampled image, respectively. For the audio modality, we extract the spectrogram of size $H_a\times W_a\times 1$ from the audio stream, where $H_a$ and $W_a$ denote the spatial dimensions of the spectrogram. Due to the mismatch between the channel dimension of audio and visual modalities, the number of channels of the audio spectrogram is inflated from $1$ to $3$ by simple repetition to ensure same input dimensions for the shared ViT architecture. As per the convention of ViT \cite{50650}, we decompose the RGB image into $M$ non-overlapping patches, which is given by $
\boldsymbol V = {v^1,v^2,v^3,.....v^M}$, where $v^i$ denote individual patches of the visual modality. Similarly, spectrogram of the audio stream is converted to $N$ patches as $
\boldsymbol A = {a^1,a^2,a^3,.....a^N}$. 

\subsection{Model Architecture}
The unique aspect of the proposed model architecture is that the full set of parameters of the ViT pretrained on ImageNet is being shared across the audio and visual modalities. This feature provides a unified framework to process both audio and visual streams using a modality-agnostic shared encoder, while being computationally-efficient by reducing the number of parameters. For masked data modeling, we use joint encoders and joint decoders to reconstruct tokens similar to \cite{gong2023contrastive}. In particular, the feature representations of the audio and visual tokens obtained from the shared ViT backbone is concatenated and fed to the joint encoders and joint decoders. The joint encoders and joint decoders are implemented using two layers of self-attention blocks and six layers of self-attention decoder blocks respectively. 


\subsection{Self-Supervised Training}
ViTs have been shown to be promising in effectively capturing self-supervised representations based on contrastive learning and masked data modeling as they are found to complement each other \cite{park2023what}. In this work, We employ a dual self-supervised learning approach that integrates contrastive learning and autoencoding objectives while incorporating asymmetric masking of audio and visual tokens. To leverage the benefits of coordinated and joint feature representations, we use contrastive learning on the coordinated representations and masked data modeling on the joint feature representations \cite{gong2023contrastive}. Contrastive learning across audio and visual tokens help to effectively capture the cross-modal correspondences, while masked data modeling retains the modality-specific information, which can be beneficial for several downstream tasks. 
To further improve the cross-modal relationships across the audio and visual tokens, we introduce asymmetric masking into the self-supervised learning framework.      

\subsubsection{Asymmetric Masking}
In most prior works 
\cite{gong2023contrastive}, fixed masking ratio has been used for both audio and visual tokens. Instead, we employ asymmetric masking by using different masking ratios for audio and visual modalities. The proposed asymmetric masking helps to better capture the cross-modal relationships by learning a diverse number of tokens across audio and visual modalities, which leads to obtain robust audio-visual feature representations. More specifically, we chose different masking ratios for audio and visual modalities by randomly assigning a masking ratio between 30\% to 60\% in each iteration. The masking ratio is uniformly sampled across audio and visual modalities over multiple iterations to ensure that the audio and visual modalities are masked in equal proportions.

\subsubsection{Audio-Visual Contrastive Loss}
The natural pairing of audio and visual information in videos provides useful information that can act as self-supervision to extract audio-visual feature representations. After obtaining the tokenized inputs of audio $
\boldsymbol A$ and visual $\boldsymbol V$ data, asymmetric masking is performed and the unmasked tokens are fed to the shared ViT backbone to obtain the feature representations of the unmasked audio and visual tokens. Now, mean pooling is done on the feature representations of the unmasked tokens of each modality to obtain global representation for the image and spectrogram for visual ($\boldsymbol F_v$) and audio modality ($\boldsymbol F_a$) respectively.   
Given the audio and visual feature representations of the corresponding images and spectrograms, the contrastive loss is obtained as 
\begin{equation}
    L_c(X_a, X_v) = -\frac1B \sum_{i=1}^{B}log\frac{exp(f(\boldsymbol F_a,\boldsymbol F_v)/\tau)}{\sum_{j=1}^{N}exp(f(\boldsymbol F_a,\boldsymbol F_v)/\tau)}
\end{equation}
where $F_a$ and $F_v$ denotes features obtained from mean pooling of unmasked tokens of audio and visual modalities, $f$ denotes cosine similarity function, $\tau$ represents temperature, and $B$ is the batch size.
\subsubsection{Masked Data Modeling Loss}
Masked autoencoders is one of the promising lines of research paradigms for self-supervised learning of ViTs in various vision tasks \cite{9879206}. The basic idea is to mask random patches of the input image and the unmasked tokens are fed to the encoder. Then the features of unmasked tokens along with the masked tokens are fed to the decoder to reconstruct the masked tokens. By masking higher proportion of tokens, it helps to reduce the computational overhead, while exhibiting strong performance \cite{zhang2022how}. Recently, masked autoencoders has been extended to the context of audio-visual learning by reconstructing the audio and visual tokens from the latent representation space \cite{gong2023contrastive,10377052}. More specifically, the feature representations of the unmasked audio and visual tokens are added with the modality type embeddings (to specify the modality in the joint embedding) along with positional embeddings and concatenated to obtain the joint embeddings of the audio and visual tokens similar to \cite{gong2023contrastive}. Now the trainable masked tokens along with modality-type embeddings are added to the joint embeddings and fed to the joint decoder to reconstruct the masked audio and visual tokens. We use mean square error loss to penalize the reconstructed tokens with the original tokens which is given by  
\begin{equation}
    L_r = -\frac1B \sum_{i=1}^{B}{(A - \Tilde{A}) + (V - \Tilde{V})  }
\end{equation}
where $\Tilde{A}$ and $\Tilde{V}$ are the reconstructed tokens of audio and visual modalities.

\begin{table*}
  \caption{Performance of the proposed approach with various architectures on the validation and Vox1-O sets.}\label{tab1}
  \centering
   \begin{tabular}{llllllll}
    \toprule
   \textbf{Audio } & \textbf{Visual} &  \textbf{Total} &\multicolumn{2}{c}{\textbf{Validation Set}}  & \multicolumn{2}{c}{\textbf{Vox1-O Set}} \\
 \textbf{Encoder} &  \textbf{Encoder} & \textbf{Parameters $\downarrow$} &\textbf{EER $\downarrow$} &  \textbf{minDCF $\downarrow$} & \textbf{EER $\downarrow$} &  \textbf{minDCF $\downarrow$} \\
    \midrule
AST-B & ViT-B & 164M & 8.217 & 0.654 & 8.915 & 0.723 \\ 
ViT-B & ViT-B & 200M & 7.043 & 0.549 & 7.298 & 0.615\\ 
\multicolumn{2}{c}{ViT-B (shared) } &  100M & 6.921 & 0.513 & 6.998 & 0.572 \\ 
\multicolumn{2}{c}{ViT-L (shared) } &  332M & \textbf{6.135} & \textbf{0.472}  & \textbf{6.344} & \textbf{0.487}\\ 
    \bottomrule
  \end{tabular}  
\end{table*}

\begin{table}
\footnotesize
  \caption{Performance of the proposed approach in comparison to state-of-the-art models on the validation and Vox1-O sets.}\label{tab2}
  \centering
   \begin{tabular}{lllll}
    \toprule
   \textbf{Fusion} &   \multicolumn{2}{c}{\textbf{Validation Set}}  & \multicolumn{2}{c}{\textbf{Vox1-O Set}} \\
\textbf{Method} &  \textbf{EER $\downarrow$} &  \textbf{minDCF $\downarrow$} & \textbf{EER $\downarrow$} &  \textbf{minDCF $\downarrow$} \\
    \midrule
Deep Cleanse \cite{10096814} &  2.476 &  0.203 & 2.409 & 0.198\\ 
 JCA \cite{praveen2023audiovisual} & 2.173  & 0.126 & 2.214 & 0.129 \\ 
 DCA \cite{praveen2024dynamic} & 2.138  & 0.119 & 2.172 & 0.121 \\
 RJCA \cite{praveen2024audiovisual} & 1.851 & 0.112 & 1.975 & 0.116 \\ \hline
Ours (audio only) & 7.349 & 0.569 & 7.482 & 0.573 \\
Ours (visual only) & 7.983 & 0.625 & 8.159 & 0.617 \\
Ours (audio-visual) & \textbf{6.135} & \textbf{0.472} & \textbf{6.344} & \textbf{0.487} \\
    \bottomrule
  \end{tabular}  
\end{table}

\begin{table}
\footnotesize
  \caption{Impact of individual components on the performance of the proposed approach on the validation and Vox1-O sets.}\label{tab3}
  \centering
   \begin{tabular}{lllll}
    \toprule
   \textbf{Method} &   \multicolumn{2}{c}{\textbf{Validation Set}}  & \multicolumn{2}{c}{\textbf{Vox1-O Set}} \\
 &  \textbf{EER $\downarrow$} &  \textbf{minDCF $\downarrow$} & \textbf{EER $\downarrow$} &  \textbf{minDCF $\downarrow$} \\
    \midrule
w/ CL only & 6.012 & 0.459 & 6.298 & 0.481  \\ 
w/ CL and MDM & \textbf{6.135} & \textbf{0.472} & \textbf{6.344} & \textbf{0.487} \\ \hline
w/ Symm. masking  & 6.181 & 0.496 & 6.362 & 0.493 \\ 
w/ Asymm. masking  & \textbf{6.135} & \textbf{0.472} & \textbf{6.344} & \textbf{0.487}  \\ 
    \bottomrule
  \end{tabular}  
\end{table}
\subsection{Overview of the proposed framework}
After obtaining the individual loss components of the CL loss and MDM loss, the final training objective is given by 

\begin{equation}
    L = L_r + \lambda L_c 
\end{equation}
where $\lambda$ denotes scaling factor for contrastive loss component. 

After training, we discard joint encoders and joint decoders and use the shared ViT backbone for obtaining the embeddings of audio and visual modalities. During testing, we feed the audio and visual inputs to the trained ViT backbone to obtain the corresponding embeddings, followed by average of audio and visual embeddings to obtain audio-visual embedding. Since we perform averaging of audio and visual embeddings, the dimensionality of the audiovisual embedding remains same as that of the individual embeddings, thereby enabling to handle even missing modalities at the time of testing.
\section{Results and Discussion}

\subsection{Datasets}
The proposed approach has been evaluated on Voxceleb1 dataset \cite{Nagrani17}, obtained from Youtube videos under challenging environments without labels. The dataset consists of 148,642 videos of 1251 speakers across a wide range of ethnicities, accents, professions, and ages, which is gender balanced with 55\% of speakers being male. The duration of each video ranges from 4 to 145 seconds. Following 
\cite{praveen2023audiovisual,praveen2024audiovisual}, we split the development set of 1211 speakers to 1150 speakers for training and 61 speakers for validation. The results are reported on both validation set and test set of 40 speakers with 37720 trials. Performance evaluation is done using Equal Error Rate (EER) and minimum Detection Cost Function (minDCF).  
\subsection{Ablation Studies}
To better understand the performance of the proposed approach, we have performed a series of experiments with different architectures, as shown in Table \ref{tab1}. First, we used an Audio Spectrogram Transformer (AST) \cite{gong21b_interspeech} for the audio modality and ViT-B \cite{50650} for visual modality. Next, we replace the AST architecture with ViT-B for audio modality and retain the same for visual modality, which showed improved performance. To understand the impact of unified framework, we share the full set of parameters of ViT-B architecture for both audio and visual modalities, resulting in slight improvement in performance. This shows that unified framework helps to better capture the audio-visual correspondences in a shared representation space, while training with less number of parameters. The proposed unified approach by sharing the full set of parameters has been evaluated with two different architectures with different number of parameters to understand the impact of scalability. By increasing the size of the architecture, we can observe that the performance has been improved, demonstrating that vision transformers are scalable audio-visual learners. 

We also performed experiments to analyze the impact of the training objectives of Contrastive Learning (CL) loss and Masked Data Modeling (MDM) loss. In both of these experiments, we used asymmetric masking. We can observe that contrastive learning plays a crucial role in capturing the audio-visual correspondences. Furthermore, integrating MDM loss provides a marginal improvement with a small increase in the number of parameters due to the joint encoders and joint decoders. Next, we implemented the proposed approach with the conventional symmetric masking, followed by the proposed asymmetric masking strategy. We can observe that the asymmetric masking enforces to better capture the audio-visual relationships in the contrastive learning framework.  

\subsection{Comparison to state-of-the-art}
The proposed approach has been compared with the state-of-the-art audiovisual methods with the same experimental protocol on Voxceleb1 dataset. The only difference is that the proposed approach is trained without using labels in a self-supervised learning framework whereas other methods are trained in a fully supervised setting. 
We have also evaluated the performance of the proposed approach using the individual audio and visual modalities along with audio-visual framework to evaluate the impact of missing modalities. Note that the model has been trained on audio-visual inputs, while the scenarios of missing modalities is evaluated only during testing. First, we consider the case of evaluating only with the audio modality with missing visual modality, followed by only with visual with missing audio modality. We can observe that the proposed unified framework can robustly handle missing modalities and exhibits comparable performance.

\section{Conclusion}
In this work, we have explored the prospect of developing a unified framework for audio-visual speaker verification in a self-supervised setting. We have shown the potential of ViTs to effectively capture robust audio-visual feature representations, while maintaining low computational complexity, enhancing scalability of the model. We further introduced asymmetric masking to enforce disproportionate masking of audio and visual tokens to better capture the cross-modal correspondences. Moreover, the proposed unified framework can effectively handle audio, visual or audio-visual inputs, offering more flexibility in training and testing. Extensive experiments are conducted on Voxceleb1 dataset to demonstrate the competitive performance of the proposed approach.


\bibliographystyle{IEEEtran}
\bibliography{mybib}

\begin{thebibliography}{10}
\providecommand{\url}[1]{#1}
\csname url@samestyle\endcsname
\providecommand{\newblock}{\relax}
\providecommand{\bibinfo}[2]{#2}
\providecommand{\BIBentrySTDinterwordspacing}{\spaceskip=0pt\relax}
\providecommand{\BIBentryALTinterwordstretchfactor}{4}
\providecommand{\BIBentryALTinterwordspacing}{\spaceskip=\fontdimen2\font plus
\BIBentryALTinterwordstretchfactor\fontdimen3\font minus \fontdimen4\font\relax}
\providecommand{\BIBforeignlanguage}[2]{{%
\expandafter\ifx\csname l@#1\endcsname\relax
\typeout{** WARNING: IEEEtran.bst: No hyphenation pattern has been}%
\typeout{** loaded for the language `#1'. Using the pattern for}%
\typeout{** the default language instead.}%
\else
\language=\csname l@#1\endcsname
\fi
#2}}
\providecommand{\BIBdecl}{\relax}
\BIBdecl

\bibitem{WANG2021215}
M.~Wang and W.~Deng, ``Deep face recognition: A survey,'' \emph{Neurocomputing}, vol. 429, pp. 215--244, 2021.

\bibitem{Desplanques2020}
B.~Desplanques, J.~Thienpondt, and K.~Demuynck, ``{ECAPA-TDNN: Emphasized Channel Attention, Propagation and Aggregation in TDNN Based Speaker Verification},'' in \emph{Proc. Interspeech}, 2020, pp. 3830--3834.

\bibitem{liou24_interspeech}
S.-H. Liou, P.-C. Chan, C.-P. Chen, T.-C. Lin, C.-L. Lu, Y.-H. Cheng, H.-F. Chuang, and W.-Y. Chen, ``Enhancing ecapa-tdnn with feature processing module and attention mechanism for speaker verification,'' in \emph{Interspeech}, 2024, pp. 2120--2124.

\bibitem{7298682}
F.~Schroff, D.~Kalenichenko, and J.~Philbin, ``Facenet: A unified embedding for face recognition and clustering,'' in \emph{CVPR}, 2015, pp. 815--823.

\bibitem{8953658}
J.~Deng, J.~Guo, N.~Xue, and S.~Zafeiriou, ``Arcface: Additive angular margin loss for deep face recognition,'' in \emph{CVPR}, 2019, pp. 4685--4694.

\bibitem{8682749}
R.~Li, N.~Li, D.~Tuo, M.~Yu, D.~Su, and D.~Yu, ``Boundary discriminative large margin cosine loss for text-independent speaker verification,'' in \emph{IEEE ICASSP}, 2019, pp. 6321--6325.

\bibitem{10.1007/978-3-031-19778-9_7}
A.~Bulat, S.~Cheng, J.~Yang, A.~Garbett, E.~Sanchez, and G.~Tzimiropoulos, ``Pre-training strategies and datasets for facial representation learning,'' 2022, p. 107–125.

\bibitem{10096659}
D.~Cai, W.~Wang, M.~Li, R.~Xia, and C.~Huang, ``Pretraining conformer with asr for speaker verification,'' in \emph{IEEE ICASSP}, 2023.

\bibitem{8683477}
S.~Shon, T.-H. Oh, and J.~Glass, ``Noise-tolerant audio-visual online person verification using an attention-based neural network fusion,'' in \emph{IEEE ICASSP}, 2019, pp. 3995--3999.

\bibitem{9350195}
Y.~Qian, Z.~Chen, and S.~Wang, ``Audio-visual deep neural network for robust person verification,'' \emph{TASLP}, vol.~29, 2021.

\bibitem{251181}
Seyed, C.~Greenberg, E.~Singer, D.~Olson, L.~Mason, and J.~Hernandez-Cordero, ``The 2019 nist audio-visual speaker recognition evaluation.''\hskip 1em plus 0.5em minus 0.4em\relax Odyssey Workshop, 2020-05-18 2020.

\bibitem{sun23_interspeech}
P.~Sun, S.~Zhang, Z.~Liu, Y.~Yuan, T.~Zhang, H.~Zhang, and P.~Hu, ``{A Method of Audio-Visual Person Verification by Mining Connections between Time Series},'' in \emph{Proc. INTERSPEECH}, 2023.

\bibitem{praveen2023audiovisual}
G.~P. Rajasekhar and J.~Alam, ``Audio-visual speaker verification via joint cross-attention,'' in \emph{Speech and Computer:25th International Conference, SPECOM}, 2023, p. 18–31.

\bibitem{10559458}
J.~Gui, T.~Chen, J.~Zhang, Q.~Cao, Z.~Sun, H.~Luo, and D.~Tao, ``A survey on self-supervised learning: Algorithms, applications, and future trends,'' \emph{IEEE Transactions on PAMI}, vol.~46, no.~12, pp. 9052--9071, 2024.

\bibitem{kim24c_interspeech}
J.~ho~Kim, H.-S. Heo, B.-J. Lee, Y.~Kwon, M.~Lee, and H.-J. Yu, ``Self-supervised speaker verification with relational mask prediction,'' in \emph{Interspeech}, 2024, pp. 2655--2659.

\bibitem{fathan24_interspeech}
A.~Fathan, X.~Zhu, and J.~Alam, ``On the impact of several regularization techniques on label noise robustness of self-supervised speaker verification systems,'' in \emph{Interspeech}, 2024, pp. 2670--2674.

\bibitem{9964072}
Y.~Gong, A.~H. Liu, A.~Rouditchenko, and J.~Glass, ``Uavm: Towards unifying audio and visual models,'' \emph{IEEE Signal Processing Letters}, vol.~29, pp. 2437--2441, 2022.

\bibitem{haliassos2024unified}
A.~Haliassos, R.~Mira, H.~Chen, Z.~Landgraf, S.~Petridis, and M.~Pantic, ``Unified speech recognition: A single model for auditory, visual, and audiovisual inputs,'' \emph{arXiv preprint arXiv:2411.02256}, 2024.

\bibitem{LAVISH_CVPR2023}
Y.-B. Lin, Y.-L. Sung, J.~Lei, M.~Bansal, and G.~Bertasius, ``Vision transformers are parameter-efficient audio-visual learners,'' in \emph{IEEE CVPR}, 2023.

\bibitem{10.1007/978-3-031-72630-9_18}
Y.-B. Lin and G.~Bertasius, ``Siamese vision transformers are scalable audio-visual learners,'' in \emph{Computer Vision -- ECCV}, 2025, pp. 303--321.

\bibitem{gong2023contrastive}
Y.~Gong, A.~Rouditchenko, A.~H. Liu, D.~Harwath, L.~Karlinsky, H.~Kuehne, and J.~R. Glass, ``Contrastive audio-visual masked autoencoder,'' in \emph{ICLR}, 2023.

\bibitem{wen2021seeking}
P.~Wen, Q.~Xu, Y.~Jiang, Z.~Yang, Y.~He, and Q.~Huang, ``Seeking the shape of sound: An adaptive framework for learning voice-face association,'' in \emph{IEEE/CVF Conference on Computer Vision and Pattern Recognition}, 2021.

\bibitem{Nagrani18c}
A.~Nagrani, S.~Albanie, and A.~Zisserman, ``Learnable pins: Cross-modal embeddings for person identity,'' in \emph{Proc. of ECCV}, 2018.

\bibitem{DBLP:conf/iclr/WenILRS19}
Y.~Wen, M.~A. Ismail, W.~Liu, B.~Raj, and R.~Singh, ``Disjoint mapping network for cross-modal matching of voices and faces,'' in \emph{ICLR}, 2019.

\bibitem{9414260}
L.~Sarı, K.~Singh, J.~Zhou, L.~Torresani, N.~Singhal, and Y.~Saraf, ``A multi-view approach to audio-visual speaker verification,'' in \emph{IEEE ICASSP}, 2021, pp. 6194--6198.

\bibitem{Tao2020}
R.~Tao, R.~K. Das, and H.~Li, ``{Audio-Visual Speaker Recognition with a Cross-Modal Discriminative Network},'' in \emph{Proc. Interspeech}, 2020, pp. 2242--2246.

\bibitem{zhu2022unsupervised}
B.~Zhu, K.~Xu, C.~Wang, Z.~Qin, T.~Sun, H.~Wang, and Y.~Peng, ``Unsupervised voice-face representation learning by cross-modal prototype contrast,'' in \emph{{Proceedings of IJCAI}}, 7 2022, pp. 3787--3794.

\bibitem{selvakumar24_interspeech}
A.~Selvakumar and H.~Fashandi, ``Getting more for less: Using weak labels and av-mixup for robust audio-visual speaker verification,'' in \emph{Interspeech 2024}, 2024, pp. 4728--4732.

\bibitem{10096814}
R.~Tao, K.~A. Lee, Z.~Shi, and H.~Li, ``Speaker recognition with two-step multi-modal deep cleansing,'' in \emph{IEEE ICASSP}, 2023, pp. 1--5.

\bibitem{praveen2024dynamic}
R.~G. Praveen and J.~Alam, ``Dynamic cross attention for audio-visual person verification,'' \emph{arXiv preprint}, 2024.

\bibitem{praveen2024audiovisual}
------, ``Audio-visual person verification based on recursive fusion of joint cross-attention,'' \emph{arXiv preprint}, 2024.

\bibitem{gnanapraveen24_odyssey}
R.~{Gnana Praveen} and J.~Alam, ``Cross-modal transformers for audio-visual person verification,'' in \emph{Odyssey Workshop}, 2024, pp. 240--246.

\bibitem{9741340}
D.~Cai, W.~Wang, and M.~Li, ``Incorporating visual information in audio based self-supervised speaker recognition,'' \emph{IEEE/ACM Transactions on ASLP}, vol.~30, pp. 1422--1435, 2022.

\bibitem{10096925}
H.~Chen, H.~Zhang, L.~Wang, K.~A. Lee, M.~Liu, and J.~Dang, ``Self-supervised audio-visual speaker representation with co-meta learning,'' in \emph{IEEE ICASSP}, 2023, pp. 1--5.

\bibitem{10314722}
B.~Han, Z.~Chen, and Y.~Qian, ``Self-supervised learning with cluster-aware-dino for high-performance robust speaker verification,'' \emph{IEEE/ACM Transactions on ASLP}, vol.~32, pp. 529--541, 2024.

\bibitem{9157330}
J.~Deng, J.~Guo, E.~Ververas, I.~Kotsia, and S.~Zafeiriou, ``Retinaface: Single-shot multi-level face localisation in the wild,'' in \emph{2020 IEEE/CVF Conference on CVPR}, 2020, pp. 5202--5211.

\bibitem{50650}
A.~Dosovitskiy, L.~Beyer, A.~Kolesnikov, D.~Weissenborn, X.~Zhai, T.~Unterthiner, M.~Dehghani, M.~Minderer, G.~Heigold, S.~Gelly, J.~Uszkoreit, and N.~Houlsby, ``An image is worth 16x16 words: Transformers for image recognition at scale,'' in \emph{ICLR}, 2021.

\bibitem{park2023what}
\BIBentryALTinterwordspacing
N.~Park, W.~Kim, B.~Heo, T.~Kim, and S.~Yun, ``What do self-supervised vision transformers learn?'' in \emph{ICLR}, 2023. [Online]. Available: \url{https://openreview.net/forum?id=azCKuYyS74}
\BIBentrySTDinterwordspacing

\bibitem{9879206}
K.~He, X.~Chen, S.~Xie, Y.~Li, P.~Dollár, and R.~Girshick, ``Masked autoencoders are scalable vision learners,'' in \emph{2022 IEEE/CVF Conference on CVPR}, 2022, pp. 15\,979--15\,988.

\bibitem{zhang2022how}
Q.~Zhang, Y.~Wang, and Y.~Wang, ``How mask matters: Towards theoretical understandings of masked autoencoders,'' in \emph{Advances in Neural Information Processing Systems}, 2022.

\bibitem{10377052}
M.-I. Georgescu, E.~Fonseca, R.~T. Ionescu, M.~Lucic, C.~Schmid, and A.~Arnab, ``Audiovisual masked autoencoders,'' in \emph{2023 IEEE/CVF ICCV}, 2023, pp. 16\,098--16\,108.

\bibitem{Nagrani17}
A.~Nagrani, J.~S. Chung, and A.~Zisserman, ``Voxceleb: a large-scale speaker identification dataset,'' in \emph{INTERSPEECH}, 2017.

\bibitem{gong21b_interspeech}
Y.~Gong, Y.-A. Chung, and J.~Glass, ``{AST: Audio Spectrogram Transformer},'' in \emph{Interspeech}, 2021, pp. 571--575.

\end{thebibliography}

\end{document}